\documentclass[11pt,a4paper]{article}
\usepackage[hyperref]{emnlp2020}
\usepackage{times}
\usepackage{latexsym}
\usepackage{multirow}
\usepackage{fixltx2e}
\usepackage{tikz}
\usepackage{graphicx}
\usepackage{caption}
\usepackage{sidecap}
\usepackage{subcaption}
\usetikzlibrary{bayesnet}
\usetikzlibrary{arrows}
\usetikzlibrary{fit,positioning}
\usetikzlibrary{arrows, automata, positioning, quotes}

\usepackage{amsmath,amsfonts,bm}
\def\mBertDis{\emph{BERT-Disc }}
\def\mBertLog{\emph{BERT-LogLP }}
\def\mBertExp{\emph{BERT-DExp }}
\def\mBertFlow{\emph{BERT-FlowLP }}
\def\mBertGMM{\emph{BERT-GMM }}

\def\mBertExpEND{\emph{BERT-DExp}}

\def\mGruDis{\emph{BiGRU-Disc }}
\def\mGruLog{\emph{BiGRU-LogLP }}
\def\mGruExp{\emph{BiGRU-DExp }}
\def\mGruFlow{\emph{BiGRU-FlowLP }}

\def\mDis{\emph{Disc }}
\def\mLog{\emph{LogLP }}
\def\mExp{\emph{DExp }}
\def\mFlow{\emph{FlowLP }}
\def\mGMM{\emph{GMM }}

\def\mLogEND{\emph{LogLP}}
\def\mExpEND{\emph{DExp}}
\def\mFlowEND{\emph{FlowLP}}
\def\mGMMEND{\emph{GMM}}

\def\num{y}
\def\anomaly{\tilde{y}}
\def\nGMMmix{{\em{K }}}
\def\nGMMmixEND{{\em{K}}}
\def\NUMMASK{\texttt{[\#MASK] }}
\def\NUMMASKEND{\texttt{[\#MASK]}}
\def\nRange{[1, 1e^{16}]}
\def\tEmbTokens{$e^\textsc{Tok}$ }

\def\tEmbPositions{$e^\textsc{Pos}$ }
\def\tEmbValues{$e^\textsc{Num}$ }
\def\tEmbValuesEND{$e^\textsc{Num}$}
\def\tEmbValuesAtI{$e_i^\textsc{Num}$ }
\def\tEmbValuesAtIEND{$e_i^\textsc{Num}$}
\def\embDim{H}

\def\tDec{$g_\theta$ }

\def\tInv{$g^{-1}_\theta$ }
\def\tInvEND{$g^{-1}_\theta$}

\def\enc{f_\gamma }
\def\encEND{f_\gamma}
\def\encBert{{\em BERT }}
\def\encGru{{\em BiGRU }}
\def\encBertEND{{\em BERT}}
\def\encGruEND{{\em BiGRU}}
\def\encX{\enc(\sequence)}

\def\inputDigit{{\em{emb\textsubscript{dig} }}}
\def\inputExp{{\em{emb\textsubscript{exp} }}}
\def\inputDigitEND{{\em{emb\textsubscript{dig}}}}

\def\sequence{\mX}
\def\latent{z}
\def\condYX{P(y|\sequence)}

\def\fullcondYX{P_{\theta,\gamma}(y|\sequence)}

\def\dataNews{{\textbf{FinNews }}}
\def\dataSci{{\textbf{Sci }}}
\def\dataDollar{{\textbf{FinNews-\$ }}}
\def\dataNewsEND{{\textbf{FinNews}}}
\def\dataSciEND{{\textbf{Sci}}}

\def\dataAll{{\em{All }}}

\def\mExpBert{{\em{ExpBert }}}

\def\metricAUC{\emph{AUC} }	
\def\metricAUCEND{\emph{AUC}}

\def\metricSAUC{\emph{S-AUC} }	
\def\metricLMAE{\emph{LMAE }}
\def\metricExpAcc{\emph{E-Acc }}

\def\metricRAUCEND{\emph{R-AUC}}
\def\metricSAUCEND{\emph{S-AUC}}		
\def\metricLMAEEND{\emph{LMAE}}
\def\metricExpAccEND{\emph{E-Acc}}

\def\taskMNM{MNM }
\def\taskMNMEND{MNM}
\def\taskNAD{NAD }
\def\taskNADEND{NAD}

\def\tnRange{[$1, 1e^{16}$]}

\def\pred{\hat{y}}
\def\tpred{$\hat{y}$ }

\def\vsigma{{\bm{\sigma}}}

\newcommand{\tb}[1]{\textbf{#1}}

\def\eqref#1{equation~\ref{#1}}

\def\floor#1{\lfloor #1 \rfloor}
\def\1{\bm{1}}
\newcommand{\train}{\mathcal{D}}
\newcommand{\valid}{\mathcal{D_{\mathrm{valid}}}}
\newcommand{\test}{\mathcal{D_{\mathrm{test}}}}

\def\vmu{{\bm{\mu}}}

\def\mX{{\bm{X}}}

\DeclareMathAlphabet{\mathsfit}{\encodingdefault}{\sfdefault}{m}{sl}
\SetMathAlphabet{\mathsfit}{bold}{\encodingdefault}{\sfdefault}{bx}{n}

\def\sR{{\mathbb{R}}}

\DeclareMathOperator*{\argmax}{arg\,max}

\usepackage{array}
\newcolumntype{H}{>{\setbox0=\hbox\bgroup}c<{\egroup}@{}}

\usepackage{microtype}

\aclfinalcopy 

\title{An Empirical Investigation of Contextualized Number Prediction}

\author{Daniel Spokoyny \\
  Carnegie Mellon University\\
  \texttt{dspokoyn@cs.cmu.edu} \\\And
  Taylor Berg-Kirkpatrick \\
  UC San Diego \\
  \texttt{tberg@ucsd.edu} \\}

\date{}

\begin{document}
\maketitle
\begin{abstract}
We conduct a large scale empirical investigation of contextualized number prediction in running text.
Specifically, we consider two tasks: (1) \textit{masked number prediction} -- predicting a missing numerical value within a sentence, and (2) \textit{numerical anomaly detection} -- detecting an errorful numeric value within a sentence. We experiment with novel combinations of contextual encoders and output distributions over the real number line. Specifically, we introduce a suite of output distribution parameterizations that incorporate latent variables to add expressivity and better fit the natural distribution of numeric values in running text, and combine them with both recurrent and transformer-based encoder architectures. We evaluate these models on two numeric datasets in the financial and scientific domain.
Our findings show that output distributions that incorporate discrete latent variables and allow for multiple modes outperform simple flow-based counterparts on all datasets, yielding more accurate numerical prediction and anomaly detection. We also show that our models effectively utilize textual context and benefit from general-purpose unsupervised pretraining.
\end{abstract}\label{abstract}
\section{Introduction}
Pretraining large neural architectures (e.g. transformers \cite{devlin-etal-2019-bert,2019t5}) on vast amounts of unlabeled data has lead to great improvements on a variety of NLP tasks.
Typically, such models are trained using a masked language modeling (MLM) objective and the resulting contextualized representations are finetuned for a particular downstream task like question answering or sentence classification \cite{devlin-etal-2019-bert,Lan2020ALBERTAL}. In this paper, we focus on a related modeling paradigm, but a different task. Specifically, we investigate contextualized number prediction: predicting a real numeric value from its textual context using an MLM-style modeling objective. We conduct experiments on two specific variants: (1) \textit{masked number prediction} (\taskMNMEND), in which the goal is to predict the value of a masked number token in a sentence, and (2) \textit{numerical anomaly detection} (\taskNADEND), with the goal of deciding whether a specific numeric value in a sentence is errorful or anomalous. 
In contrast with more standard MLM training setups, here we specifically care about the accuracy of the trained masked conditional distributions rather than the contextualized representations they induce.
While successful models for these tasks are themselves useful in applications like typo correction and forgery detection \cite{chen-etal-2019-numeracy}, better models of numeracy are essential for further improving downstream tasks like question answering, numerical information extraction \cite{mirza-etal-2017-cardinal,saha-etal-2017-bootstrapping} or numerical fact checking \cite{thorne-vlachos-2017-extensible}, as well as for processing number-heavy domains like financial news, technical specifications, and scientific articles. 
Further, systems that detect anomalous numbers in text have applications in practical domains -- for example, medicine  \citep{thimbleby2010reducing} -- where identification of numerical entry errors is critical.

\begin{figure*}
\centering
  \includegraphics[scale=0.35]{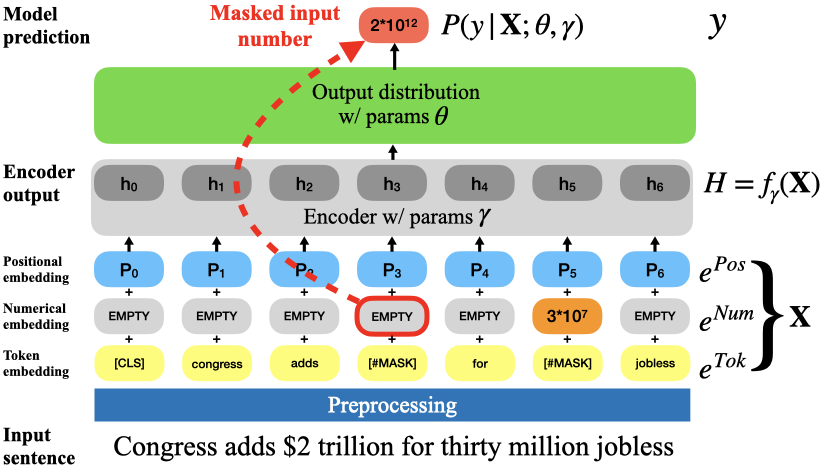}
    \caption{
  Outline of our model architecture consisting of a sentence representation $\sequence$ which is fed to the encoder with parameters $\gamma$ and an output distribution over the real number line with parameters $\theta$.
  In this example our masked numerical objective is to predict the masked out ``2 trillion'' quantity $\num$.
  Note that our model is able to use a numerical embedding of the unmasked input $3*10^7$ value (``thirty million'') as part of the context.}
  \label{fig:example}
\end{figure*}

Our modeling approach to contextualized number prediction combines two lines of past work. First, following \citet{chen-etal-2019-numeracy}, we treat number prediction as a sentence-level MLM problem where only numerical quantities are masked. However, \citet{chen-etal-2019-numeracy} focused on predicting the discrete exponent of masked numbers as a classification problem. In contrast, \citet{spithourakis2018numeracy} demonstrate the utility of predicting full numerical quantities in text, represented as real numbers, but do so in a language modeling framework, conditioned only on left context. Here, we propose a novel setup that combines full-context encoding (i.e.~both left and right contexts) with real-valued output distributions for modeling numerical quantities in text. In Figure~\ref{fig:example}, we illustrate an example where we aim to predict  ``2 trillion'' as a quantity on the real number line. 

We expand upon past work by conducting a large scale empirical investigation that seeks to answer three questions: (1) Which encoding strategies yield more effective representations for numbers in surrounding context?
(2) Which encoding architectures provide the best representations of surrounding context?
(3) What are the most effective real-valued output distributions to model masked number quantities in text? To answer these questions, we propose a suite of novel real-valued output distributions that add flexibility through the use of learned transformation functions and discrete latent variables. We conduct experiments for both \taskMNM and \taskNAD tasks on two large datasets in different domains, combining output distributions with both recurrent and transformer-based encoder architectures, as well as different numeric token encoding schemes. Further, while
\citet{chen-etal-2019-numeracy} studied a specific type of \taskNAD (detecting exaggerated numbers in financial comments), we examine several \taskNAD variants with different types of synthetic anomalies that are found to arise in practice across different domains of data. Finally, we further compare results with a strong discriminative baseline.

\section{Models}
Our goal is to predict numbers in their textual contexts. The way we approach this is similar to masked language modeling (MLM), but instead of masking and predicting all token types, we only mask and predict tokens that represent numeric values. For example in Figure~\ref{fig:example} we wish to predict that the value of the masked number \NUMMASK should be $2 \times 10^{12} \in \sR$ given the surrounding context.

For notational simplicity, we describe our model as predicting a single missing numeric value in a single sentence.
However, like other MLMs (see section \ref{section:experiments}), during training we will mask and predict multiple numeric values simultaneously.
Let $\sequence$ be a sentence consisting of $N$ tokens where the $k$th token is a missing numerical value, $\num$.
The goal of our model is to predict the value of $\num$ conditioned on $\sequence$.
We will use common notation for from similar setups and simply treat the $k$th token in $X$ as a masked numeric value, \NUMMASKEND.

Our models $\fullcondYX$ consist of three main components: an input representation of the sentence, a contextual encoder with parameters $\gamma$ which summarizes the sentence, and an output distribution with parameters $\theta$ over the real number line.
In this section we will describe our strategies for numerical input representation, the two types of contextual encoders we use, along with different formulations of numerical output distributions.

\subsection{Input Context Representation}\label{section:embeddings}
We first describe the input representation for the textual context $\sequence$ that will be passed into our model's encoder.
We let $x_i$ represent the $i$th token in the input sequence.
Like related MLMs that leverage transformers (which is one type of encoder we consider in experiments) we separate the representation of $x_i$ into several types of embeddings.
We include a positional embedding \tEmbPositions and a word-piece token embedding \tEmbTokens like the original BERT.
We also introduce our new numeric value embedding \tEmbValues to help us learn better numerical representations.
Finally, as shown in Figure~\ref{fig:example}, the input representation for token $x_i$ is the sum of these three \embDim-dimensional embeddings.

If the token at position $i$ represents a numerical quantity, we replace it with a special symbol \NUMMASKEND, and represent its numerical value using \tEmbValuesAtIEND.\footnote{We exclude segment type embeddings since we do not perform next sentence prediction. We also found it helpful to use the zero vector as the numerical embedding for \tEmbValuesAtI if position i is not a quantity.}
We use the extraction rules detailed in Section~\ref{section:preprocessing} to find the numbers in our input sequence.
In the next section we will describe two strategies for numerical representation \tEmbValuesEND.

\subsubsection{Digit-RNN Embedding}
The large range ($\nRange$ in our data) of numerical values prevents them from being used directly as inputs to neural network models as this results in optimization problems due to the different scales of parameters.
One strategy to learn embeddings of numerical values has been shown by \citet{saxton2018analysing} which used character-based RNNs to perform arithmetic operations such as addition and multiplication.
We conduct experiments with a similar strategy and represent each number in scientific notation (d.ddde+d) with 6 digits of precision as a string.
We then use a digit-RNN to encode the string and use the last output as \tEmbValuesEND.

\subsubsection{Exponent Embedding}
A simpler approach to represent numbers would be to explicitly learn embeddings for their magnitudes.
Magnitudes have been shown to be a key component of the internal numerical representation of humans and animals \cite{ansari2016number,whalen1999nonverbal, Dehaene1998AbstractRO}.
We conduct experiments with an encoding scheme that learns embeddings for base-10 exponents.

\subsection{Context Encoder}\label{encoders}
The encoder's goal is to summarize the surrounding text, along with other numbers that appear therein.
We define $\bm{H} = \encX$ where the encoder $\enc$ is a function of the context $\sequence$, and $\bm{H}$ is the hidden representation of the encoder's last layer.
Next, we describe two encoder architectures: a transformer and a recurrent approach.

\subsubsection{Transformer Encoder}\label{encoders:transformer}
Transformer architectures pretrained on vast amounts of data have led to breakthroughs in textual representation learning \cite{Yang2019XLNetGA,liu2019roberta,Lan2020ALBERTAL,2019t5}.
We use the 12-layer BERT-base architecture \cite{devlin-etal-2019-bert} with the implementation provided by Huggingface \cite{Wolf2019HuggingFacesTS}.
We use the original BERT's word-piece vocabulary with 30,000 tokens and add a new \NUMMASK token.

\subsubsection{BiGru Encoder}\label{encoders:bigru}
Previous methods focusing on the related task of predicting the order of magnitude of a missing number in text showed that RNNs were strong models for this task \citep{chen-etal-2019-numeracy}.
In our real-valued output task we use a bidirectional Gated Recurrent Unit (\encGruEND), the best performing model from \citet{chen-etal-2019-numeracy}.
We use a one-layer BiGRU with a 64-dimensional hidden state and a dropout layer with a 0.3 dropout rate.
We use the same pretrained word-piece embeddings from BERT as this allows us to directly compare the two encoders.

\subsection{Real-valued Output Distributions}
\begin{figure*}
  	\centering
  	\includegraphics[width=.9\linewidth]{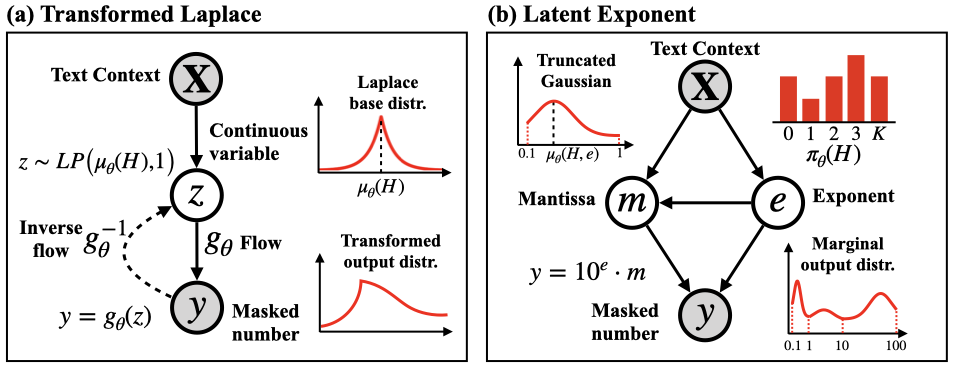}
  	\caption{
  		{\em Left (a):} We depict our \mLog and \mFlow graphical models along with the latent and output distributions.
  		{\em Right (b):} Probabilistic graphical model of our latent \mExp model.}\label{fig:genflow}
\end{figure*}

In early experiments, we observed that simple continuous distributions (e.g. Gaussian or Laplace) performed poorly.
Since numbers can have ambiguous or underspecified units, and further, since numbers in text are heavy-tailed, asymmetric or multi-modal output distributions may be desirable.
For this reason, we propose several more flexible output distributions, some which include learned transforms and others which include latent variables (both well-known methods for adding capacity to real-valued distributions), to parameterize $\condYX$.

\subsubsection{Log Laplace}
A common method for constructing expressive probability density functions is to pass a simple density through a transformation (e.g. a flow or invertible mapping function).
As an initial example (and our first output distribution), we describe the log Laplace distribution as a type of flow. Since numbers in text are not distributed evenly on the number line due to a long tail of high magnitudes, a simple trick is to instead model the log of numeric values. If the base distribution is Laplace, this yields a log Laplace distribution, which we describe next as an exponential transformation.

In Figure~\ref{fig:genflow}, we illustrate our \mLog model with a continuous intermediate variable $\latent$, encoder $\encEND$, with $exp$ as the transformation, \tDec, and consequently $log$ as \tInvEND.
In equation~\ref{eq-log} we show our generative process and training objective where both \tDec and \tInv are deterministic functions with no parameters.
We let $\mu_\theta(\bm{H})$ denote a single layer MLP that outputs the location parameter of the base Laplace distribution on $z$, which is transformed to produce the output variable, $y$. More precisely:
\begin{equation}\label{eq-log}
\vspace{-5pt}
\includegraphics[width=.8\linewidth]{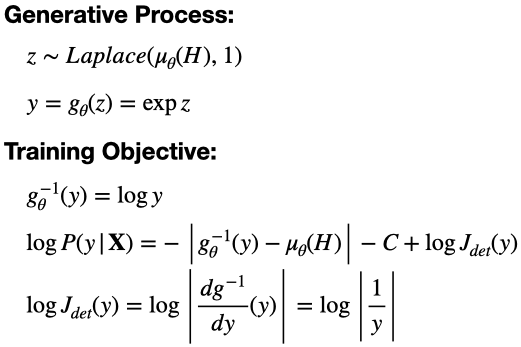}
\end{equation}

\vspace{-5pt}
\subsubsection{Flow-transformed Laplace}
The $\exp$ transformation may not be the ideal choice for our data. For this reason we consider a parameterized transform (flow) to add further capacity to the model. For our purposes, we are restricted to 1-dimensional transformations $g : \mathbb{R} \rightarrow \mathbb{R}$.
Further, by restricting the class of functions, we ensure an efficient way of computing the log-derivative of the inverse flow, which allows us to efficiently compute likelihood. We conduct experiments with the simple parameterized flow described in Equation~\ref{eq-flow}.
We use a single layer MLP to independently predict each parameter \emph{a,b,c} from $\bm{H}$, the output of $\encX$.
We also scale the range of $b,c$ to be between [0.1, 10] using a Sigmoid activation.
Similarly to the \mLog setting, $\mu_\theta(\bm{H})$ is a single layer MLP which predicts the location parameter of the Laplace.

\begin{equation}\label{eq-flow}
\includegraphics[width=.8\linewidth]{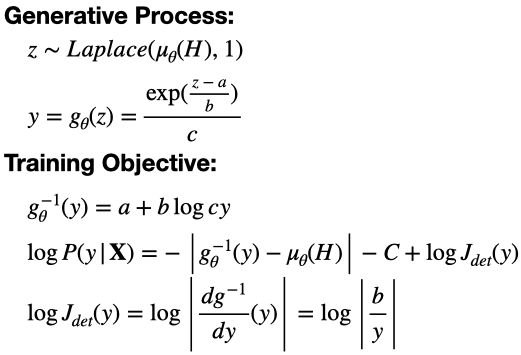}
\end{equation}
\noindent This parameterization of flow is designed to allow for (1) re-centering of the input variable (via parameter $a$), (2) re-scaling of the input (via parameter $b$), and (3) re-scaling of the output (via parameter $c$). Together, this leads to a family of inverse flows that are all log-shaped (i.e. they compress higher values), yet have some flexibility to change intercept and range.


\subsubsection{Discrete Latent Exponent}
While \mFlow adds flexibility over the \mLog model, both have the drawback of only being able to produce unimodal output distributions.\footnote{In principle, more complicated flows could also have multiple modes -- though they are more challenging to construct and optimize.}
A well-established approach to parameterizing multi-modal densities is to use a mixture model.
The mixture component is determined by a discrete latent variable in contrast with the continuous intermediate variable introduced in the flow-based models.
In Figure~\ref{fig:genflow} we show our \mExp model where $e$ represents an exponent sampled from a multinomial distribution, and $m$ is the mantissa sampled from a truncated Gaussian.

Prior work has shown the effectiveness of cross-entropy losses on numerical training \cite{saxton2018analysing,chen-etal-2019-numeracy}.
For this reason we use a truncated Gaussian on the range of [0.1,1] to generate $m$, which effectively restricts back-propagation to a single mixture component for a given observation.
The combination of exponent and mantissa prediction allows us to benefit from the effectiveness of cross-entropy losses, while at the same time getting more fine-grained signal from the mantissa loss.
In Equation~\ref{eq-exp} we show the \mExp generative process and training objective.
We let $\pi_\theta(\bm{H})$ denote a single layer MLP that outputs the multinomial parameters of $P(e|X)$.
Similarly, we let $\mu_\theta(\bm{H},e)$ denote a two layer MLP with a [.1,1] scaled Sigmoid that outputs the mean parameter of the mantissa normal distribution.
\begin{equation}\label{eq-exp}
\vspace{-5pt}
\includegraphics[width=.8\linewidth]{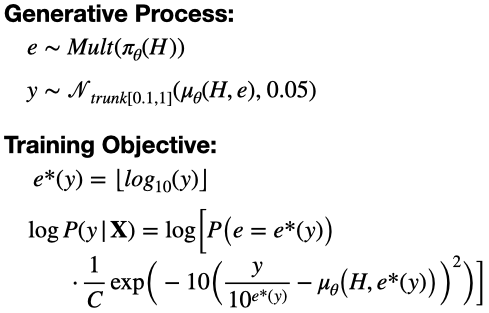}
\vspace{-5pt}
\end{equation}

\subsubsection{Gaussian Mixture Model}\label{section:GMM}
Inspired by the best performing model from \citet{spithourakis2018numeracy} we also compare with a Gaussian mixture model (\mGMMEND).
This model assumes that numbers are sampled from a weighted mixture of \nGMMmix independent Gaussians. 
During training the mixture from which a particular point was sampled from is not observed and so it is treated as a latent variable.
We can optimize the marginal log-likelihood objective by summing over the $K$ mixtures.
In equation \ref{eq-gmm}, \mGMM has \nGMMmix mixtures parameterized by $K$ means and variances $\vmu, \vsigma$, respectively.
Following \citet{spithourakis2018numeracy}, we pre-train the parameters $\vmu, \vsigma$ on all the numbers in our training data $\train$ using EM.
The means and variances are then fixed and our masked number prediction model only predicts mixture weights during training and inference.
We let $\pi_\theta(\bm{H})$ denote a single layer MLP that outputs the mixture weights $P(e|X)$.

\begin{equation}\label{eq-gmm}
\includegraphics[width=.8\linewidth]{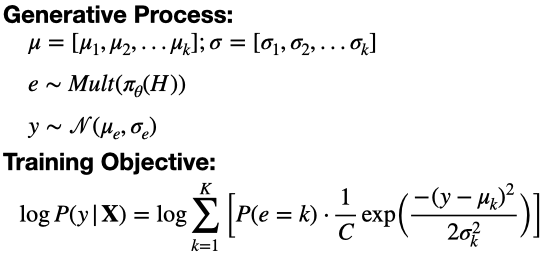}
\end{equation}
\section{Data}
\paragraph{Financial news} Financial news documents are filled with many different ratios, quantities and percentages which make this domain an ideal test-bed for \taskMNMEND.
The \dataNews is a collection of 306,065 financial news and blog articles from websites like Reuters\footnote{www.kaggle.com/jeet2016/us-financial-news-articles}.
We randomly break the documents into [train, valid, test] splits with [246065, 30000, 30000] respectively.

Since \dataNews has many occurrences of dates and years, we also evaluate on a subset corpus, \dataDollar, to measure effectiveness at modeling only dollar quantities in text. \dataDollar is constructed exactly as \dataNews, with the added requirement that the number is preceded by a dollar sign token (\$). For all training and testing on \dataDollar, we only predict dollar values. 

\paragraph{Academic papers} Academic papers have diverse semantic quantities and measurements that make them an interesting challenge numeracy modeling. For this reason, we also use S2ORC, a newly constructed dataset of academic papers \cite{lo-wang-2020-s2orc}. We use the first 24,000 full text articles, randomly splitting into [20000, 2000, 2000] [train, valid, test] splits.
\footnote{We also filter articles from only these categories \{Geology, Medicine, Biology, Chemistry, Engineering, Physics, Computer science, Materials science, Economics, Business, Environmental science\}.}
We refer to this dataset as \dataSciEND.
All three datasets follow the same preprocessing discussed below and summary statistics are provided in Table \ref{table-data-stats}.

\subsection{Preprocessing \label{section:preprocessing}}
Financial news, academic papers, and Wikipedia articles all have different style-guides that dictate how many digits of precision to use or whether certain quantities should be written out as words.
While such stylistic queues might aid models in better predicting masked number \emph{strings}, we are specifically focused on modeling actual numeric values for two reasons: (1) reduced dependence on stylistic features of the text domain leads to better generalization to new domains, and (2) the numerical value of a numeric token conveys its underlying meaning and provides a finer-grained learning signal.
For example currencies are usually written as a number and magnitude like {\em \$32 million} however, many quantities can be written out as cardinals {\em sixty thousand trucks}.
We normalize our input numbers so that changing the style from {\em five} to {\em 5} does not change our output predictions.

As exemplified in Figure~\ref{fig:example}, the aim of our approach is to incorporate both numbers as context and numbers as predictions (i.e.~2 trillion and thirty million in the example). 
For this reason, before tokenization we employ heuristics to combine numerals, cardinals and magnitudes into numerical values, whilst removing their string components.
We also use heuristics to change ordinals into numbers.
By following this normalization preprocessing procedure we get higher diversity of naturally occurring quantitative data and mitigate the bias towards some particular style guide.

For both \dataNews and \dataSci we lowercase the text and ignore signs $(+, -)$, so all numbers are positive and restrict magnitudes to be in \tnRange.
We discard sentences that do not have numbers or where the numbers are outside of our specified range.
We also filter out sentences that have less than eight words and break up sentences longer than 50 words.\footnote{Sentences under eight words in length tended to be titles of articles with the date as the only numeric quantity.}
We do not use the special token {\em [SEP]} and all examples are truncated to a maximum length of 128 tokens.

\begin{table*}
\centering
\begin{tabular}{l |lll|lll|lll}
\hline
& \multicolumn{3}{c}{\dataNews} & \multicolumn{3}{c}{\dataDollar} & \multicolumn{3}{c}{\dataSci} \\
& \textbf{train} & \textbf{valid} & \textbf{test} & \textbf{train} & \textbf{valid} & \textbf{test} & \textbf{train} & \textbf{valid} & \textbf{test} \\
\hline

\#instances     & 522996 & 58095    & 64433     & 188286 & 22338 & 23281    & 360514 & 36523 & 36104 \\
avg-length      & 102.5  & 108.3    & 108.9     & 115.2  & 115.4 & 116.1    & 125.6  & 126.4 & 126.5 \\
\%numbers       & 8.8    & 9.3      & 9.6       & 13.0  & 12.7  & 13.2     & 7.1   & 7.2  & 7.1  \\
\hline
min             & 1.0    & 1.0      & 1.0       & 1.0    & 1.0   & 1.0      & 1.0    & 1.0    & 1.0    \\
median 50       & 313.0  & 250.0    & 329.0     & 2016.0   & 2016.0  & 2016.0     & 9.0    & 8.0    & 9.0    \\
median 75       & 3141.0 & 2558.0   & 3500.0    & $\sim10^{4}$  & $\sim10^{4}$ & $\sim10^{4}$    & 42.0   & 40.0   & 43.0   \\
median 90       & $\sim10^{6}$  & $\sim10^{6}$    & $\sim10^{6}$     & $\sim10^{7}$  & $\sim10^{7}$ & $\sim10^{7}$    & 1959.0 & 1948.0 & 1972.1 \\
max             & $\sim10^{15}$ & $\sim10^{14}$   & $\sim10^{15}$      & $\sim10^{15}$ & $\sim10^{14}$& $\sim10^{15}$     & $\sim10^{15}$ & $\sim10^{14}$ & $\sim10^{15}$   \\
\hline
\end{tabular}
\caption{\label{table-data-stats}
Statistics on our datasets. The top half of the table reveals the number of examples per data split, the average length of sentences, and the fraction of tokens that are numbers. The bottom half shows summary statistics for number values in both datasets.}
\end{table*}


\section{Experiments} 
In this section we explain our experimental setup, starting with our evaluation metrics, implementation details, results, and ablation analyses. We use the following naming convention for models: we specify the encoder (\encGruEND, \encBertEND) first, followed by one of our four output distributions (\mLogEND, \mFlowEND, \mExpEND, \mGMMEND).

\subsection{Evaluation}
For the \taskMNM task on $\valid$ and $\test$ splits we randomly select a single number to mask out from the input and predict.
We let \tpred denote the model's $\argmax$ prediction from $\condYX$ and $\num$ as the actual observed number.
In equation \ref{eq-lmae} and \ref{eq-exp-acc} we show how we calculate log-MAE (\metricLMAEEND) and exponent accuracy (\metricExpAccEND), both of which use log base 10.

\small
\begin{equation}\label{eq-lmae}
    \begin{aligned}
    \text{\metricLMAE} = \frac{1}{|\test|}\sum_{\test}|\log\num - \log\pred| \\
    \end{aligned}
\end{equation}
\vspace{-.4cm}
\small
\begin{equation}\label{eq-exp-acc}
\text{\metricExpAcc} = \frac{1}{|\test|}\sum_{\test}=
\left\{\begin{matrix}
1 & if \floor{\log\num} = \floor{\log\pred}\\
0 & otherwise
\end{matrix}\right.
\end{equation}
\normalsize

\subsection{Numerical Anomaly Detection}
Both \metricLMAE and \metricExpAcc metrics test the model's argmax prediction and not the entire $\condYX$ distribution.
We next consider the \taskNAD task where our models need to discern the true number versus some anomaly.
We let $\anomaly$ denote an anomaly and describe two different ways, [{\em string, random}], we construct an anomalous example.
For {\em string} we use the true $\num$ and randomly perform one of three operations [{\em add, del, swap}]: inserting a new digit, deleting an existing digit, and swapping the first two digits respectively.
For {\em random}, we randomly sample a number from the training data $\train$ as our anomaly.
We choose these string functions as they constitute a large part of numerical entry errors \cite{thimbleby2010reducing,wiseman2011taxonomy}. Further, {\em random} mimics a copy-paste error.
We report the AUC of a ROC curve for both types as random-anomaly (\metricRAUCEND) and string-anomaly (\metricSAUCEND) respectively, using the model's output density to rank the true value against the anomaly.

\subsection{Implementation Details}\label{section:experiments}
We train all models with stochastic gradient descent using a batch-size of 32 for 10 epochs. We use early stopping with a patience of three on the validation loss.
For pretrained \encBert encoder experiments, we use two learning rates $\{3e^{-5}, 1e^{-2}\}$ for all pretrained parameters and newly added parameters respectively.
For all non-pretrained \encBert experiments and all \encGru encoders we use a single learning rate of $2e^{-2}$.

\citet{devlin-etal-2019-bert} propose a two step process to generate masked tokens. First, select tokens for masking with an independent probability of 15\%. Second, for a selected token: With 80\% probability replace it with a \texttt{[MASK]}, 10\% replace it with a random token, and 10\% leave it unchanged. Since there are fewer numbers than text tokens, we use a higher probability of 50\% for selection. We follow a similar strategy for masking numbers: 80\% of the time masking out the number, 10\% of the time randomly substituting it with a number from train, and 10\% of the time leaving it unchanged.

\paragraph{Baselines:} We also consider a fully discriminative baseline trained to predict real vs. fake numbers with binary cross entropy loss.
The negative numerical samples are randomly drawn from training set numbers to match exactly the random-anomaly task.
During training each positive datum has one negative example and is trained in the same batch-wise fashion. 
When this model uses exponent embeddings for output numbers, \inputExp, we can also calculate the exponent accuracy by selecting the exponent embedding with highest model score as a predicted value. We include this approach in experiments as a non-probabilistic alternative to our four output distributions.

\subsection{Results}
\begin{table*}[]
\centering
\small
\begin{tabular}{lHllll}
\hline
\textbf{Model}                        & \tb{Loss} & \tb{LMAE}$\downarrow$ &\tb{\metricExpAcc}$\uparrow$&\tb{r-AUC}$\uparrow$&\tb{s-AUC}$\uparrow$ \\
\hline
Train-Mean                            & -       & 7.69       & 1.03        & -     & -      \\ 
Train-Median                          & -       & 1.88       & 5.52        & -     & -      \\ 
\hline
\mGruDis                              & 1.37   & -           & 55.8       & 0.756 & 0.646  \\ 
\mGruLog                              & 7.63   & 0.671       & 58.8       & 0.675 & 0.548  \\ 
\mGruFlow                             & 7.11   & 0.622       & 61.8       & 0.694 & 0.591  \\ 
\mGruExp                              & 2.09   & 0.576       & 71.5       & 0.843 & 0.821  \\ 
\hline
\mBertDis                             & 1.54    & -           & 62.7        & 0.762  & 0.656  \\ 
\mBertGMM \nGMMmixEND=255             & 10.12   & 1.18        & 21.3       & 0.585 & 0.440  \\ 
\mBertLog                             & 7.39    & 0.5666      & 64.9       & 0.686 & 0.557  \\ 
\mBertFlow                            & 6.98    & 0.5732      & 65.5       & 0.717 & 0.609  \\ 
\mBertExp                             & 1.97    & {\bf 0.500} & {\bf 74.6} &{\bf 0.861} &{\bf 0.828}  \\ 
\end{tabular}
\caption{\label{tresults:fin-all}Results on \dataNews where all models use input exponent embeddings \inputExp and all \encBert encoders are pretrained. We also include the mean and median number from training $\train$ as simple baselines.}
\end{table*}
We ran all combinations of encoders and output distributions using input exponent embeddings on \dataNews and show the results in Table \ref{tresults:fin-all}.
We train the \mGMM model with four different settings of \nGMMmix $\in$ \{31, 63, 127, 255\} and report results for the highest-performing setting.

Comparing the two encoders, we find that \encBert results in stronger performance across all metrics and all output distributions.
Although both settings share the same pretrained embedding layers, the pretrained transformer architecture has higher capacity and is able to extract more relevant numerical information for both \taskMNM and \taskNADEND.

We find that the parameterized \mFlow model was generally better across all metrics under both encoders compared to the \mLog model.
With the weaker \encGru encoder, the \mLog model's \metricSAUC is only 0.04 better than random guessing.

The \mExp model was the best performing output distribution across all metrics and both encoders, yielding on average 10\% higher \metricExpAcc and a gain of 0.13 on \metricAUCEND.
This means that \mExp had the best overall fit in terms of the predicted mode ($\argmax$) as well as the overall density $\condYX$.

In contrast, \mGMM, which is also a discrete latent variable model capable of outputting a multimodal distribution,
underperformed across all metrics.
There was little effect from adjusting the number of mixture components, with slight improvements using more mixtures.
One possible reason for the \mGMM model's worse performance is that the mixtures are fit and fixed before training without any of the surrounding textual information.
Quantities such as dates and years have many textual clues, but the model's initial clustering may group them together with other quantities. 
We also found that, empirically, optimization for this model was somewhat unstable.

Finally the \mDis baseline was the second best performing model on \taskNAD, though on \taskMNM it showed worse \metricExpAcc than \mLog and \mFlow models.
This baseline benefited from being directly trained for \taskNAD, which may explain it's under-performance on \taskMNM metrics.
Due to the comparatively worse performance of both the \encGru encoder and the \mGMM output distribution, we exclude them from the remainder of our experiments.

\subsection{Ablations}\label{section:ablation}
\begin{table*}
\centering
\small
\begin{tabular}{lHll lHlH|ll}
\hline
\textbf{Ablation Type}                       &\tb{Loss} &\tb{LMAE}$\downarrow$ &\tb{\metricExpAcc}$\uparrow$ &\tb{r-AUC}$\uparrow$ & \tb{r-F1}$\uparrow$ & \tb{s-AUC}$\uparrow$ & \tb{s-F1}$\uparrow$ &\tb{all-LMAE}$\downarrow$ &\tb{all-\metricExpAcc}$\uparrow$\\
\hline
\textbf{Numerical Input Embedding}               &      &     &       &     &       &      &     &         &  \\
\mBertExp (All \#'s Masked)          & 2.18     & 0.656     & 66.5      & 0.831    & 0.787      & 0.809      & 0.762     & 0.656        & 66.5 \\
\mBertExp + \inputExp                & 1.97     & 0.500     & 74.6      & 0.861    & 0.820      & 0.828      & 0.778     & 0.888        & 62.2 \\
\mBertExp + \inputDigit              & 1.99     & 0.506     & 74.4      & 0.858    & 0.815      & 0.826      & 0.776     & 0.920        & 62.1 \\
\mBertExp + \inputExp + \inputDigit  & 1.97     & 0.498     & 74.9      & 0.861    & 0.818      & 0.828      & 0.778     & 0.899        & 62.3 \\
\hline
\textbf{No Pretraining}               &      &     &       &     &       &      &     &         &  \\
\mBertExp + \inputExp         & 2.15     & 0.615     & 68.8      & 0.840    & 0.797      & 0.810      & 0.764     & 0.889        & 60.6 \\
\mBertFlow + \inputExp        & 7.36     & 0.769     & 57.9      & 0.670    & 0.677      & 0.563      & 0.667     & 0.861        & 54.4 \\
\mBertDis + \inputExp         & 1.57     & -         & 26.9      & 0.632    & 0.685      & 0.599      & 0.679     & -            & - \\
\mBertLog + \inputExp         & 7.53     & 0.630     & 63.2      & 0.678    & 0.687      & 0.550      & 0.667     & 0.850        & 57.1 \\
\hline
\end{tabular}
\caption{\label{tresults:fin-news:input_ablations}
Ablation on \dataNews dataset. The top half of the table shows the effect of the numerical input representation. The bottom half shows performance for models trained from scratch, without leveraging pretrained BERT parameters.}
\end{table*}

\paragraph{Ablations on Numerical Embedding} We select our best performing model, \mBertExpEND, and ablate the numerical input representation on \dataNewsEND.
We compare using \inputDigitEND, \inputExp, and a version of \mExpBert which has no numerical input representation.
The top half of Table~\ref{tresults:fin-news:input_ablations} displays the results. 
We see that \inputDigit and \inputExp perform equally well.
Using no input number embeddings reduces performance by 8\% on \metricExpAcc and  0.03 \metricAUC on both anomaly metrics.
We also see that there is no benefit from combining both of these input representations, which implies that the model is able to extract similar information from each.

\paragraph{Ablations One-vs-All} To measure our model's effectiveness at using the other numbers in the input we construct an ablated evaluation \dataAll, where all input numbers are masked out.\footnote{To make comparisons exact, every test example has at least 2 numerical values so that we can perform this ablation.}
In Table~\ref{tresults:fin-news:input_ablations} we see that all models that have a numerical embedding suffer a performance drop of around 12\% \metricExpAcc and an increase of 0.4 on \metricLMAEEND.
This suggests that the model is in fact using the other quantities for its predictions.
We also find that the model with no input number embeddings does better on the \dataAll setting since it was effectively trained with fully masked input numbers.

\paragraph{Ablations on Pretraining} In the bottom half of Table~\ref{tresults:fin-news:input_ablations}, we compare the effect of starting from a pretrained transformer versus training from scratch.
We see that training from scratch hurts all models by around 6\% on \metricExpAcc and 0.02 on \metricRAUCEND.
We also note that \mBertLog seems least affected, dropping only 1\% on \metricExpAccEND.

\paragraph{Modeling Additional Domains} In this section we explore how different models behave on the alternative domain of academic papers, and how modeling is affected by focusing only dollar quantities in financial news.
In Table~\ref{tresults-fin-dol-sci-docs}, we show results for pretrained BERT encoder models with input exponent embeddings, trained and evaluated on \dataSci and \dataDollar datasets.

On the \dataSci data, the generative models have similar performance on \metricLMAE and \metricExpAcc. 
We further find that \mBertExp is still the best performing model across most metrics on both \dataSci and \dataDollar data.
The \mBertDis baseline, which is directly trained to predict anomalies, is consistently the second best across all datasets on \taskNADEND. 
Finally, we find that the \dataDollar is the most challenging of the three datasets, with \mBertExp dropping on \metricExpAcc by 20\% compared to \dataNews data.
This supports our initial reasoning that the distribution of dollar amounts is more difficult to characterize than other quantities, such as dates, which tend to cluster to smaller ranges.

\begin{table*}
\centering
\small
\begin{tabular}{l |Hrrrr|Hrrrr}
\hline
                & \multicolumn{5}{c}{\dataDollar}      & \multicolumn{5}{c}{\dataSci} \\
\tb{Model}      & \tb{Loss} & \tb{LMAE}$\downarrow$& \tb{\metricExpAcc}$\uparrow$&\tb{r-AUC}$\uparrow$& \tb{s-AUC}$\uparrow$& \tb{Loss} & \tb{LMAE}$\downarrow$ & \tb{\metricExpAcc}$\uparrow$&\tb{r-AUC}$\uparrow$& \tb{s-AUC}$\uparrow$\\
\hline
\mBertDis       & 1.07      & -          & 46.9        & 0.828        & 0.588                    & 1.37   & -           & 68.8       & 0.722        & 0.657 \\
\mBertLog       & 16.88     & 1.04       & 43.6        & 0.641        & 0.528                    & 3.66   & \textbf{0.374}       & 78.2       & 0.624        & 0.609 \\
\mBertExp       & 2.71      & {\bf 0.91} & {\bf 56.9}  & {\bf 0.867}  & {\bf 0.678}              & 1.81   & 0.385 & {\bf 81.0} & {\bf 0.786}  & {\bf 0.836} \\
\mBertFlow      & 16.29     & 1.11       & 39.3        & 0.538        & 0.518                    & 3.47   & \textbf{0.374}       & 77.6       & 0.658        & 0.672 \\
\hline
\end{tabular}
\caption{\label{tresults-fin-dol-sci-docs}
Results on \dataDollar and \dataSci where all models use input exponent embeddings \inputExp and all \encBert encoders are pretrained.
}
\end{table*}

\section{Related Work}
\paragraph{Math \& Algebraic Word Problems:}There is a wide literature on using machine learning to solve algebraic word problems \cite{ling2017program,roy2016solving,zhang2019gap}, building novel neural modules to directly learn numerical operations \cite{trask2018neural,Madsen2020Neural} and solving a variety of challenging mathematical problems \cite{saxton2018analysing,Lee2020Mathematical,Lample2020Deep}.
In these tasks, numbers can be treated as symbolic variables and computation based on these values leverages a latent tree of arithmetic operations.
This differs from our task setting since there is no ``true'' latent computation that generates all the quantities in our text given the available context.

\paragraph{Numerical Question Answering} The DROP dataset \citep{Dua2019DROP} is a new dataset that requires performing discrete numerical reasoning within a traditional question answering framework.
\citet{Andor2019GivingBA} treat DROP as a supervised classification problem, while recent work by \citet{ggb2020injecting} show how synthetic mathematical training data can build better numerical representations for DROP.
Unlike work on DROP, our primary focus is on the task of contextualized number prediction and numerical anomaly detection in text, which involve correlative predictions based on lexical context rather than concrete computation. 

\paragraph{String Embeddings} Recently, word and token embeddings have been analyzed to see if they record numerical properties (for example, magnitude or sorting order) \cite{Wallace2019Numbers,naik-etal-2019-exploring}.
This work finds evidence that common embedding approaches are unable to generalize to large numeric ranges, but that character-based embeddings fare better than the rest. However, this line of work also found mixed results on overall numeracy of existing embedding methods and further investigation is required.


\paragraph{Numerical Prediction} \citet{spithourakis2018numeracy} trained left-to-right language models for modeling quantities in text as tokens, digits, and real numbers using a \mGMMEND.
Our empirical investigation focuses on \taskMNM and considers both left and right contexts of numbers, along with a broader class of generative output distributions.
\citet{chen-etal-2019-numeracy} predict magnitudes of numbers in text and also consider a type of \taskNAD to detect numerical exaggerations on financial data.
However, this modeling approach is restricted: it can only distinguish anomalies that result in a change of exponent.
In contrast, our real-valued distributions allow us to focus on a broader suite of harder anomaly detection tasks, such as random substitutions and string input error.



\section{Conclusion}
In this work we carried out a large scale empirical investigation of masked number prediction and numerical anomaly detection in text.
We showed that using the base-10 exponent as a discrete latent variable outperformed all other competitive models.
Specifically, we found that learning the exponent representation using pretrained transformers that can incorporate left and right contexts, combined with discrete latent variable output distributions, results is the most effective way to model masked number quantities in text.
Future work might explore combining more expressive flows with discrete latent variables.

\section*{Acknowledgements}
We thank Volkan Cirik and the anonymous conference reviewers for providing valuable feedback. This project is funded in part by the NSF under
grants 1618044 and 1936155, and by the NEH
under grant HAA256044-17. The first author is supported in part by an NSF GRFP. Findings and observations do not necessarily reflect the views of funding agencies.

\bibliography{emnlp2020}
\bibliographystyle{acl_natbib}
\clearpage

\end{document}